\documentclass[conference]{IEEEtran}
\usepackage{times}
\usepackage[numbers]{natbib}
\usepackage{multicol}
\usepackage[bookmarks=true]{hyperref}
\usepackage{comment}
\usepackage{graphicx}
\usepackage{amsmath} 
\usepackage{amsthm}  
\usepackage{amssymb} 
\theoremstyle{remark}
\newtheorem{remark}{Remark}

\begin{document}

\title{Discretizing Dynamics for Maximum Likelihood Constraint Inference}

\author{
\authorblockN{Kaylene C. Stocking\authorrefmark{1},
David L. McPherson\authorrefmark{1},
Robert P. Matthew\authorrefmark{2}, and 
Claire J. Tomlin\authorrefmark{1}}
\authorblockA{\authorrefmark{1} Department of Electrical and Computer Engineering\\
University of California, Berkeley}
\authorblockA{\authorrefmark{2} Department of Physical Therapy and Rehabilitation Science\\
University of California, San Francisco}}

\maketitle

\begin{abstract}
Maximum likelihood constraint inference is a powerful technique for identifying unmodeled constraints that affect the behavior of a demonstrator acting under a known objective function. However, it was originally formulated only for discrete state-action spaces. Continuous dynamics are more useful for modeling many real-world systems of interest, including the movements of humans and robots. We present a method to generate a tabular state-action space that approximates continuous dynamics and can be used for constraint inference on demonstrations that obey the true system dynamics. We then demonstrate accurate constraint inference on nonlinear pendulum systems with 2- and 4-dimensional state spaces, and show that performance is robust to a range of hyperparameters. The demonstrations are not required to be fully optimal with respect to the objective, and the most likely constraints can be identified even when demonstrations cover only a small portion of the state space. For these reasons, the proposed approach may be especially useful for inferring constraints on human demonstrators, which has important applications in human-robot interaction and biomechanical medicine.

\end{abstract}

\IEEEpeerreviewmaketitle

\section{Introduction}
Inverse reinforcement learning (IRL) allows an agent to infer the goals driving someone's behavior and learn to complete the same task simply by observing. This is a powerful paradigm for learning new behaviors from scratch, but doesn't encompass all of the useful information we may extract from observations. For example, consider the case where the agent already has a good policy for some task, but notices that an expert demonstrator is deviating from the optimal behavior. A reasonable explanation would be that the demonstrator is acting under new environmental constraints that the agent is unaware of. As a concrete example, we can imagine a scenario where an autonomous vehicle (the agent) is following a car that suddenly swerves (the demonstrator). Since both vehicles have the same policy of avoiding collisions and following the road, the agent can infer that an obstacle suddenly appeared in the road and take evasive action even before it can detect the obstacle directly. Taking cues from other agents is an important aspect of intelligent behavior that can help compensate for problems such as sensor failure or perceptual error.

\begin{figure}[t]
    \centering
    \includegraphics[width=0.9\linewidth]{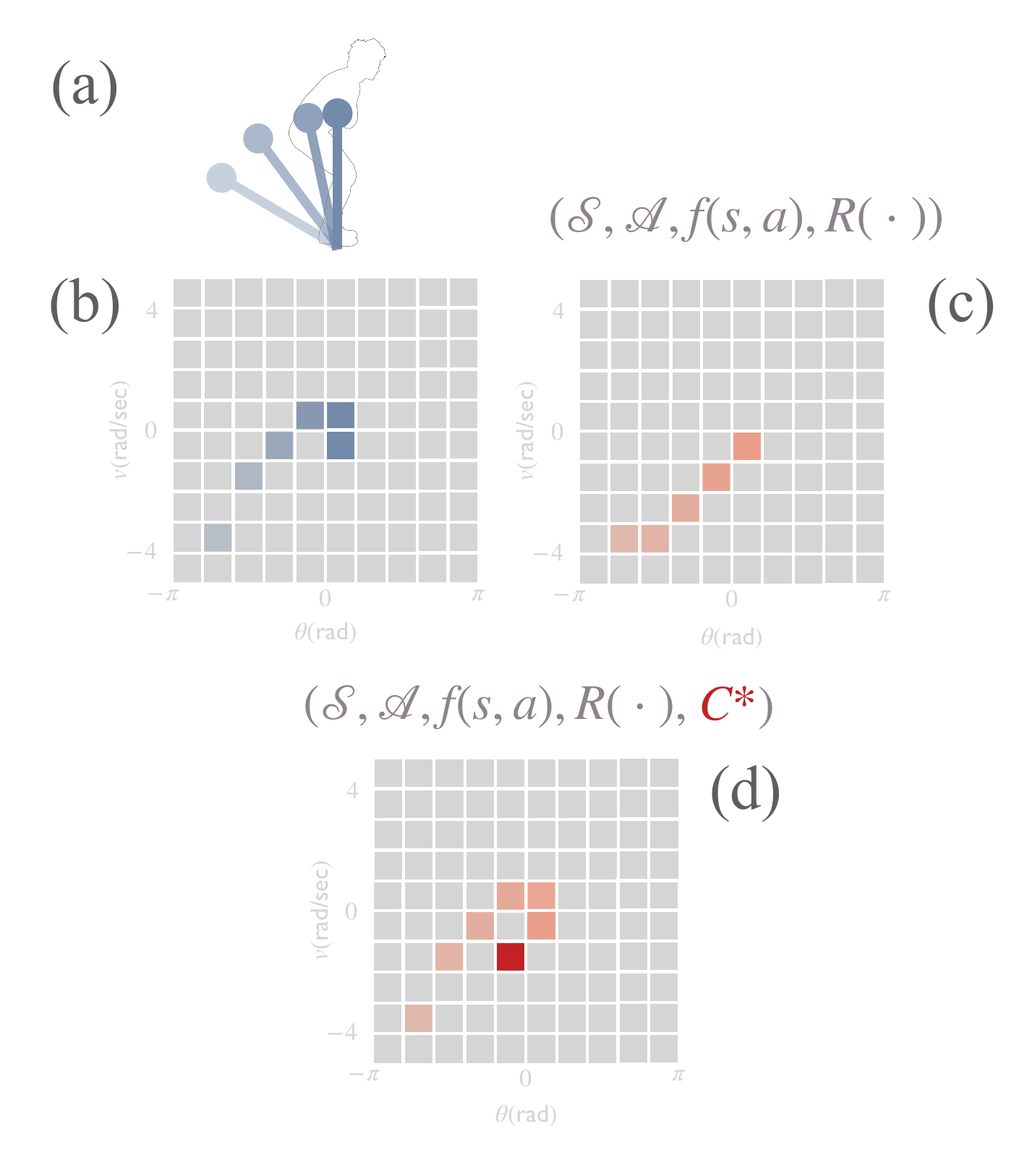}
    \caption{An overview of maximum likelihood constraint inference on a continuous system. (a) A human demonstrates task performance while obeying an unknown constraint. (b) Their continuous trajectory is projected onto the approximate state space $\mathcal{S}$ of a tabular Markov Decision Process (MDP). (c) The human's behavior deviates from what the MDP expects due to the unmodeled constraint. (d) The most likely constraint $\mathcal{C}^*$ can be inferred, causing the tabular MDP trajectory to more closely match that of the human. The inferred constraint region can be mapped back into the continuous state space for applications such as identifying potential biomechanical problems.}
    \label{fig:headlineIllustration}
\end{figure}

The process of detecting constraints that help explain the behavior of a demonstrator is called constraint inference. \citet{scobee_maximum_2019} applied the maximum entropy IRL framework to this problem, resulting in an algorithm that can identify the most likely constraints from a hypothesis set. However, this approach is limited to systems with tabular state-action spaces. This precludes its use in many real systems of interest whose dynamics are inherently continuous. In this paper, we describe a procedure for creating a tabular approximation of an arbitrary continuous system, and show that maximum likelihood constraint inference (MLCI) can be used to infer constraints on the approximated system that transfer well to the original continuous system. We analyze the effects of various approximation hyperparameters on the accuracy of constraint inference on an example 2-dimensional pendulum system. Although this analysis does not necessarily generalize to other sets of dynamics, following a similar procedure on a system of interest can indicate whether the approximation is sufficient for meaningful constraint inference. We also present a technique for estimating confidence in the inferred constraint in the form of a Bayesian probability update. 

In addition to applications where the agent wishes to use learned constraints to improve its own policy, this extension of MLCI allows us to perform constraint inference directly from observed human movements. One exciting potential application of this work is in individuals with \emph{non-specific} low-back pain, pain that is not immediately attributable to a specific pathology. This affects 84\% of people in their lifetime, with around 12\% of people being disabled from this pain \cite{balague_non-specific_2012}. Our proposed constraint inference approach would enable an explanatory biomechanical tool to infer joint level limitations from a series of full-body movements, where traditional biomechanical methods have seen limited success \cite{papi_is_2018}. This motivates the telescoping inverted pendulum model (section VI) which has been used to model different standing patterns in clinical populations \cite{papa_telescopic_1999}.

We first present related work in section II, before briefly introducing MLCI in section III and our method for translating continuous dynamics into a tabular Markov Decision Process that can be used with MLCI in section IV. We then perform experimental analysis in section V, and finally show an example with clinically motivated 4D telescoping inverted pendulum dynamics in section VI.

\section{Related Work}

Previous work on constraint inference can be split into two categories: approaches that infer the most likely constraints but require a tabular state-action space, and those that admit continuous dynamics but drop the maximum likelihood feature. In the former category, in addition to \citet{scobee_maximum_2019}, \citet{vazquez-chanlatte_learning_2018} learn task specifications, which can be thought of as a generalization of state-space constraints to include complex multi-step behaviors. Unfortunately, neither of these approaches can be applied directly to many real-world systems that are inherently continuous.

There are many proposed methods for identifying constraints in systems that cannot be tabulated. Some use heuristics such as assuming that constrained behaviors will have high intra-demonstration variance and low inter-demonstration variance, or that a maintaining an end effector in the same orientation throughout a demonstration suggests a constraint \cite{pais_learning_2013,li_learning_2017}. \cite{lin_learning_2015} presents a kinematics-based approach for learning constraints that affect how a nominal policy is executed in different environments, but doesn't assume an objective function and therefore requires demonstrations to cover much of the state space for the inference to be well-defined. \cite{mehr_inferring_2016} is specialized for online constraint inference in the context of shared autonomy, where mis-identified constraints can be corrected by the user. \cite{chou_learning_2020-1} provides a flexible approach for learning state-space constraints by sampling from possible trajectories with lower costs than the demonstrations. 

Although the present work introduces error by estimating continuous dynamics with a finite state-action space, it provides two key advantages over previous methods that work with continuous dynamics. First, using the maximum entropy framework allows us to model the demonstrators as soft-optimal with respect to a reward function, which may be especially appropriate for human demonstrators. Second, we are able to estimate and rank the most likely constraints even in situations where demonstrations cover only a small portion of the state space and do not provide enough information to fully resolve ambiguity in possible constraints.

\section{Markov Decision Processes and Maximum Likelihood Constraint Inference}

To perform maximum likelihood constraint inference (MLCI), we adapt the approach developed by \cite{scobee_maximum_2019}. In this section, we present a brief overview of the MLCI algorithm.

\subsection{Markov Decision Dynamics}
MLCI is formulated as an operation on a tabular Markov Decision Process (MDP). The MDP is a tuple of four elements:

\begin{itemize}
  \item A state space $\mathcal{S}$ to navigate. $\mathcal{S}$ is a finite set of discrete state values: $$\mathcal{S} = \{s^1, s^2, ..., s^M\}$$
  \item A set of actions $\mathcal{A}$ to decide between. $\mathcal{A}$ is a finite set of discrete input values: $$\mathcal{A} = \{a^1, a^2, \cdots, a^W\}$$
  \item A transition kernel $$P_{a_{t-1}}(s_t | s_{t-1}) : \mathcal{S} \times \mathcal{A} \times \mathcal{S} \rightarrow [0,1]$$ that determines the influence of $a_{t-1} \in \mathcal{A}$ on $s_t$. The repeated action of this transition kernel generates a sequence of states over a time horizon $t \in [0:T]$ given a sequence of action choices up to the horizon. The couple of state sequence and action sequence is the trajectory $\xi : [0:T] \rightarrow A \times S$ and the space of all possible trajectories is $\Xi$.
  \item An objective metric $R(\xi)$ that measures the quality of trajectories. $$R : \Xi \rightarrow \mathbb{R}$$
\end{itemize}

This work focuses on deterministic dynamics, so the transition kernel will be singleton distributions with zero probability of all next states except the deterministic successor $s_t = f(a_{t-1}, s_{t-1})$. That is, we focus on MDP's with transitions of the form:

\begin{align}
  P_{a_{t-1}}(s_t | s_{t-1}) =
  \begin{cases}
    1, \text{ if } s_t = f(a_{t-1}, s_{t-1}) \\
    0, \text{ otherwise}
  \end{cases}
\end{align}

MLCI requires that $\mathcal{S}$ and $\mathcal{A}$ be finite sets, and we refer to an MDP which satisfies this property as tabular.

\subsection{Maximum Entropy Likelihood on Trajectories}
This work leverages the maximum entropy likelihood distribution advanced in \cite{ziebart_modeling_2010} and extended to constraint inference in \cite{scobee_maximum_2019}.
This distribution's randomness reflects epistemological uncertainty in the estimated reward function $R(\xi)$ of the demonstrator.
Under this distribution, the likelihood of a trajectory $\xi$ is defined on the deterministic MDP as:

\begin{align}
  P(\xi) =
  \frac{
    e^{
      R(
        \xi
      )
    }
  }{Z}
\end{align}

where $Z$ is the normalizing constant:

$$
  Z = \sum_{
    \xi \in \Xi
  }
    e^{
      R(
        \xi
      )
    }
$$

This work investigates how dynamic agents avoid certain sets of states $C \subset \mathcal{S}$.
These constrained states further refine the choice distribution by zeroing out illegal choices:

\begin{align}
  P_C(\xi) =
  \begin{cases}
    \frac{
      e^{
        R(
          \xi
        )
      }
    }{Z_C},
    \text{ if } s_{t} \notin C \quad \forall t \\
    0, \text{ otherwise}
  \end{cases}
  \label{eq:BoltzmannDef}
\end{align}

Where the partition constant $Z$ decreases to $Z_C$ for this new distribution that constrains out much of the previous support.
Let $\Xi_C \subset \Xi$ be the subset of trajectories that don't violate the constraint $C$:

$$
  \Xi_C = \{
      \xi
      |
      s_t \notin C \quad \forall t
    \}
$$

So that $Z_C$ may be simply defined as:

$$
  Z_C = \sum_{
    \xi \in \Xi_C
  }
    e^{
      R(
        \xi
      )
    }
$$

\subsection{Constraint Inference}
The distribution in equation (\ref{eq:BoltzmannDef}) describes the likelihood of observing any demonstrated trajectory $\xi$ given a constraint set $C$.
Given a set of independent and identically distributed sample trajectories $\tilde{\xi} = \{\xi^{(j)} \text{ for } j \in [1,2, \cdots, N]\}$, the likelihood of observing this dataset is:

\begin{align}
  P_C(\tilde{\xi}) &= \Pi_{j=1}^{N} P(\xi^{(j)}) \\
  &= \Pi_{j=1}^{N}
  \frac{
    e^{
      R(
        \xi^{(j)}
      )
    }
  }{Z_C}
  \mathbb{I}[\xi^{(j)} \in \Xi_C]
  \\
  &=
  \frac{1}{Z_C}
  \Pi_{j=1}^{N}
  e^{
    R(
      \xi^{(j)}
    )
  }
  \mathbb{I}[\xi^{(j)} \in \Xi_C]
\end{align}

Adding a constraint to the model that helps explain the demonstrations will increase this likelihood. Therefore, the most likely constraint $C^*$ is the one that maximizes $P_{C^*}(\tilde{\xi})$. Note two properties that will aid in finding $C^*$:

\begin{remark}
  The optimal constraint set $C^*$ must have all $\xi^{(j)}$ inside of its corresponding $\Xi_{C^*}$. Otherwise its likelihood would be 0 -- a lower likelihood even than having no constraints at all. This would contradict its being the optimum.
\end{remark}

Therefore for any feasible candidate constraints, the indicator $\mathbb{I}[\xi^{(j)} \in \Xi_C]$ will always evaluate to 1. With the zero-case ruled out, the likelihood can be straightforwardly characterized by factoring out the remaining $C$-dependent component:

\begin{remark}
  When comparing the likelihood amongst feasible constraint sets,
  they are only re-scalings of the same dataset-determined constant 
  $
  \Pi_{j=1}^{N}
  e^{
    R(
      \xi^{(j)}
    )
  }
  $ by $1/{Z_C}$.
  So the maximum likelihood constraint set $C^*$ is simply whichever set $C$, amongst the feasible constraint sets, has the smallest $Z_C$.
\end{remark}

For every hypothesized constraint set $C_i$, $Z_{C_i}$ can be computed by a Bellman backup or by forward simulation.
The latter approach is favored by 
\cite{scobee_maximum_2019} as it makes a direct parallel to the seminal Maximum Entropy IRL work \cite{ziebart_modeling_2010}. Let $C_0 \subset \mathcal{S}$ be some baseline set of known constraints (e.g. the empty set for the unconstrained case).
The forward simulation relies on the fact that $Z_{C_i}$ is proportional to $P_{C^0}(\xi \in \Xi_{C_i})$.

\begin{align*}
  Z_{C_i} &= 
  \sum_{\xi \in \Xi_{C_i}}
    e^{
      R(
        \xi
      )
    }
  \\ &=
  Z_{C_0}
  \sum_{\xi \in \Xi_{C_i}}
  \frac{
    e^{
      R(
        \xi
      )
    }
  }{
    Z_{C_0}
  }
  \\ &=
  Z_{C_0} P_{C^0}(\xi \in \Xi_{C_i})
\end{align*}

It can be calculated by forward simulating the state distribution under $C_0$'s maximum entropy distribution and observing the probability that trajectories violate the constraint $C_i$ up to time $t \in [0:T]$. Call that quantity $\Phi_{i,t}$, then:

\begin{align*}
  Z_{C_i} &= Z_{C_0} P_{C^0}(\xi \in \Xi_{C_i}) \\
      &= Z_{C_0} (1 - P_{C^0}(\xi \notin \Xi_{C_i})) \\
      &= Z_{C_0} (1 - \Phi_{i,T}) \\
\end{align*}

Therefore, the most likely constraint minimizes $Z_{C_i}$, or equivalently, maximizes $\Phi_{i,T}$. The quantity $\Phi_{i,T}$ will be useful in some of our subsequent analysis.

\section{Formulation of Approximate MDP}

Given an arbitrary set of continuous dynamics of the form $\dot{x} = h(x,u)$, we wish to generate an appropriate tabular state-action space that can be used with the MLCI algorithm described in section III. We will illustrate this process with a pendulum model that we return to for experimental analysis in section V.

\subsection{Running Example: Pendulum System}

The pendulum model consists of a 2-dimensional state space (angle and angular velocity). The 1-dimensional control input is the normalized torque applied at the base of the pendulum:

\begin{equation}
    \ddot{\theta} = \frac{g}{l}\cdot sin(\theta) + u
\label{eqn:pendulum_dynamics}
\end{equation}

Where the gravitational constant $g$ and the length of the pendulum $l$ are both assumed to be 1 for simplicity. The constraint hypothesis set is an evenly spaced 10-by-10 grid of non-overlapping cells that cover the state space, for a total of 100 possible constraints. (Note that any set of state space regions is acceptable as the constraint hypothesis set, including overlapping regions or ones that do not cover the whole state space, but it is typically appropriate for them to be equally sized. This is because a larger constraint region is able to ''explain away" more demonstrator sub-optimality and is therefore likely to have a larger $\Phi_T$, making it difficult to directly compare constraint regions of different sizes when choosing the most likely one.) The demonstrator wants to arrive at a particular goal state at the end of a $\tilde{T}$ = 5s period while minimizing the total squared torque and avoiding the true constraint region, $K$: 

\begin{equation}
\begin{split}
    u^* = min\int_{0}^{\tilde{T}} u(\tilde{t})^2 d\tilde{t} \\
    \mathrm{s.t.} (\theta(0), \dot{\theta}(0)) = (\theta_0, \dot{\theta}_0) \\
    (\theta(\tilde{T}), \dot{\theta}(\tilde{T})) = (\theta_T, \dot{\theta}_T) \\
    (\theta(\tilde{t}), \dot{\theta}(\tilde{t})) \notin K \quad \forall \tilde{t} \in [0, \tilde{T}]
\end{split}
\label{eqn:pendulum_objective}
\end{equation}

Where we use $\tilde{t}$ to refer to continuous time and $t$ to refer to discrete time steps.

\subsection{Forming The Tabular State-Action Space}

 First, we choose appropriate bounds for each dimension of the state space and control input, which can come from domain knowledge or observing the range of values in the demonstrations. For the pendulum system, it is natural to bound $\theta \in [0, 2\pi]$, we select the velocity bound $\dot{\theta} \in [-6, 6]$, and the control input bound $u \in [-2, 2]$ is chosen by observing that the controls used by continuous trajectories optimizing the objective in equation (\ref{eqn:pendulum_objective}) rarely exceed this range. We then grid up the continuous state space by dividing it into disjoint cells that completely cover the bounded area. A reasonable default is to use equally sized boxes. For example, we can divide the pendulum state space into 100 cells, 10 along each dimension, each encompassing a $\frac{\pi}{5}$ rad angle width and a 1.2 rad/s angular velocity range. The set of these cells is $\mathcal{S}$. Similarly, the range of possible control inputs is divided into discrete points to give $\mathcal{A}$. We use $s$ and $a$ to label the discrete states and actions, respectively. $x_{s_t}$ is the value of the continuous state at the center point of state cell $s_t$, while $u_{a_t}$ is the value of the control input associated with discrete action choice $a_t$. 
 
 \subsection{Tabular MDP Transition Kernel and Objective}
 
 To complete the tabular MDP representation, we need to determine the transition and reward associated with each $(s \in \mathcal{S}, a \in  \mathcal{A}$). For ``gridworld" environments frequently used in inverse reinforcement learning, the agent is allowed to transition to any adjacent cell. However, for arbitrary continuous dynamics, this behavior may result in trajectories that bear little resemblance to what is possible under the true dynamics. For example, consider that in the pendulum system, allowing a transition from $\theta \in [0, \frac{\pi}{5}]$ to $\theta \in [\frac{\pi}{5}, \frac{2\pi}{5}]$ is nonsensical if the current velocity is a large negative value, regardless of the control input.

To resolve this problem, we select a constant time interval $\Delta t$ that represents the amount of time that passes between state transitions in the tabular model. For each discrete $(s \in \mathcal{S}, a \in  \mathcal{A})$, we use an ODE solver to determine the trajectory that would result from starting at the center of state cell $s$, $x_s$, and applying a constant control input of $u_a$ for $\Delta t$ time. We can then determine which state cell the agent would land in at the end of this trajectory segment, which becomes the successor state $s'$. While this is sufficient for determining appropriate discrete transitions, the start and successor cells alone do not tell us which state-based constraints may have been violated while taking a particular transition. Therefore, we also keep track of which hypothesized constraints would be violated while executing the continuous trajectory underlying the discrete transition. This ensures that the agent isn't allowed to ``warp through" constraints even when $s$ and $s'$ are not adjacent cells.

Finally, we assume that the ground-truth reward for an entire trajectory can be expressed as $R = \int_{0}^{\tilde{T}}r(x(\tilde{t}), u(\tilde{t}))d\tilde{t}$ for some function $r(x,u)$ of the continuous state and control input. We estimate the tabular reward function as $\hat{R} = \sum_{t=0}^{T} r(x_{s_t},u_{a_t}) \cdot \Delta t$, where the sequence $(s_{[0:T]}, a_{[0:T]})$ is the sequence of discrete state-action pairs over the course of a trajectory on the tabular MDP.

It is worth noting that trajectories allowable under the tabular MDP described above are not necessarily feasible or safe under the true continuous dynamics. For example, starting from different points within the same cell might result in slightly different constraint violations, while we only track violations that result from starting in the center of each cell. This is acceptable for our application because we are trying to obtain estimates of general behavior that enable reasonable likelihood-based constraint inference. Similarly, there is no well-defined mapping from a particular continuous trajectory to a feasible discrete state-action sequence under the approximate tabular dynamics. Since we only handle state-based constraints, it is sufficient to determine which possible constraints a demonstration violates without trying to construct a discrete version of the trajectory. This can be done by sampling points along the trajectory to determine which constraint regions it passes though.

The primary hyperparameters that determine the final tabular MDP are the number of cells to use for each state dimension, the number of actions, and the transition time step $\Delta t$. These parameters can be tuned using domain knowledge or by running simulated experiments with known constraints to determine which model obtains the best performance. An example of these experiments and the resulting constraint inference performance for the pendulum system is described in the following section. Once an appropriate model has been selected, it can be used with any combination of reward function and demonstration set. Additionally, if the objective of the demonstrators is known in advance, MLCI can be performed on the appropriately initialized discrete MDP as a pre-computation step, and constraints can be inferred online with very little additional computation.

\section{Analysis On Pendulum System}

\begin{figure}
    \centering
    \includegraphics[scale=0.4]{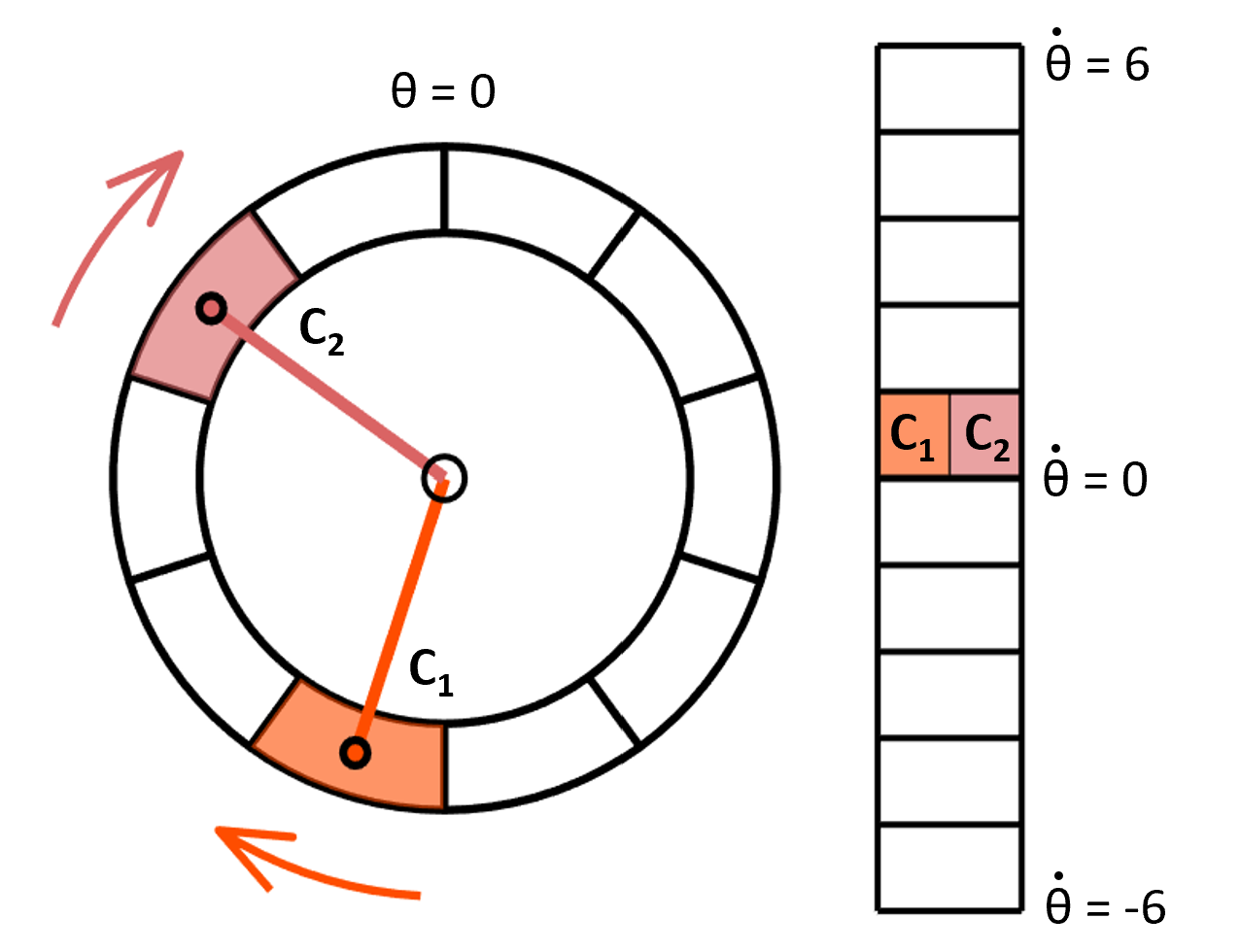}
    \caption{We test our constraint inference on a fixed-base pendulum model. The constraint hypothesis space evenly divides the state space into 100 cells, 10 along the angle axis and 10 along angular velocity. The two constraints used in our experiments, $C_1$ and $C_2$, are shown here in different shades of red. The constraints cover different angle regions but the same angular velocity.}
    \label{fig:pendulumModel}
\end{figure}

After following the procedure outlined in section IV for the pendulum system, we now have a tabular MDP representation that can be used with MLCI as described in section III. We next turn to analyzing the behavior of this approximate MDP. For our experiments, we tested two possible ground-truth constraints: $C_{1}$ prohibits $\theta \in [\pi,\frac{6}{5}\pi]$ while $\dot{\theta} \in [0,1.2]$, and $C_{2}$ prohibits $\theta \in [\frac{8}{5}\pi,\frac{9}{5}\pi]$ while $\dot{\theta} \in [0,1.2]$ Both ground-truth constraints are aligned with the constraint hypothesis set. The constraint hypothesis space is illustrated in Fig. \ref{fig:pendulumModel}. For each ground truth constraint, we randomly sampled 100 pairs of start and end states from $\Omega$ (defined in equation (\ref{eq:safeSet}) below) for agents to satisfy while optimizing the objective in equation (\ref{eqn:pendulum_objective}). Some of these start-end state pairs were ill-posed since the pendulum could not reach across them in the fixed 5 second time horizon provided. After removing these configurations, the set of demonstrations was reduced to $N = 65$ trajectories.

\subsection{Accuracy Of Tabular MDP Dynamics}

\begin{figure}
    \includegraphics[scale=0.33]{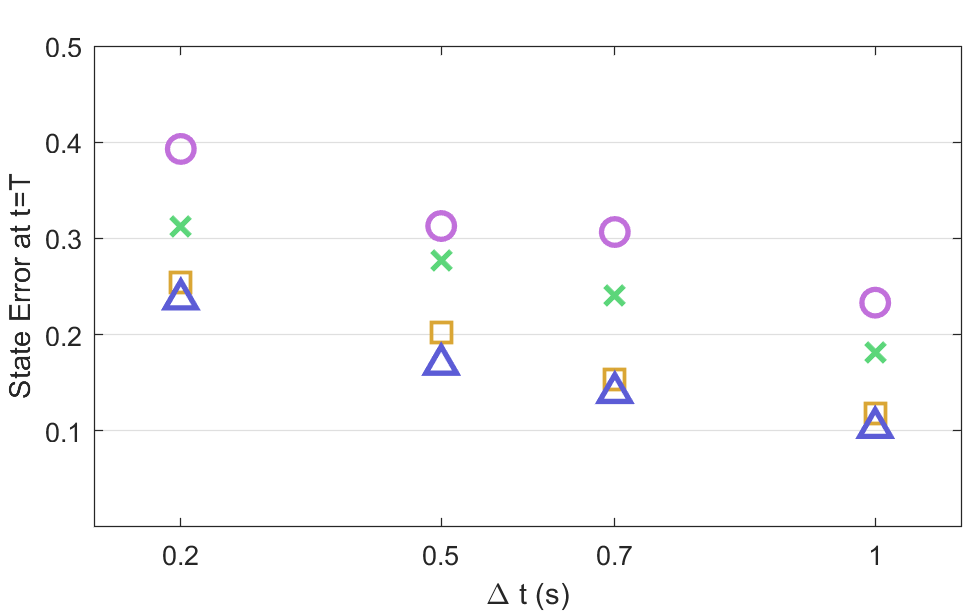}
    \caption{Dividing the state space into larger numbers of smaller cells in the MDP allows for a more accurate approximation of the continuous dynamics when executing optimal policies over 5s. A smaller $\Delta t$ means that a larger number of discrete transitions are taken over the same time horizon, leading to decreased accuracy, especially for a coarser discretized state space. Pink circles: 100 state cells, green crosses: 400, yellow squares: 900, blue triangles: 1600. Note that there is decreasing benefit to further refining the state space grid above 900 cells.}
    \label{fig:accuracy}
\end{figure}

We first examine how accurately the tabular MDP recovers the true continuous dynamics under goal-directed behavior induced by the objective function. For each ground-truth constraint and random start-goal pair, we initialized the MDP while incorporating the true constraint into the MDP dynamics (i.e., actions that would result in violating the true constraint were not allowed). We then performed a Bellman backup to determine the distribution of soft optimal policies on the tabular MDP. Intuitively, if the MDP perfectly describes the true continuous dynamics, we expect that running a simulation with the ground-truth dynamics while taking the sequence of actions determined by one of these policies will cause the agent to land exactly at the goal state. Following this intuition, we sampled and executed a random policy from each MDP and measured the normalized Euclidean distance between the final state and goal state. As shown in Fig. \ref{fig:accuracy}, increasing the number of state cells reduces the ``round-off" error associated with each discrete state transition and results in a final state that is closer to the intended goal. Since the objective function specifies a fixed time horizon, increasing $\Delta t$ decreases the number of transitions over the course of a trajectory and therefore reduces final state error as well.

\subsection{Generating Simulated Expert Demonstrations}
To understand the accuracy of constraint inference with the tabular MDP, we first need expert demonstrations that follow the ground-truth dynamics. 
For each ground-truth constraint, 100 random pairs of states were sampled to serve as the start and goal points for independent demonstrations.
These expert continuous demonstrations were synthesized using a second-order descent method with simulation time step ($\Delta t = 0.01 sec$), much finer than the $\Delta t$ used in the tabular MDP.
The demonstrations are optimized using a Gauss-Newton-style descent method known as Iterative Linear-Quadratic Regulators (or iLQR) \cite{li_iterative_2004}.
The optimization is halted after ten iterations.
For each start-goal pair, the best-of-three optimizations is picked (each with randomly sampled controls initialization) to reject optimizations that get stuck in local minima.
The optimizations that could not succeed in reaching their goal were filtered out from the dataset, reducing the dataset size to $N = 65$.

The state constraints are blocked out as rectangular polytope constraints in the continuous state-space.
They are enforced using an interior-point method that supersedes any controls (as in \cite{hoffmann_decentralized_2008}) that would reach the constrained states.
This backwards-reachable set that forms the barrier-certificate \cite{prajna_safety_2004} is computed via a Hamilton-Jacobi Isaacs Partial Differential Equation \cite{mitchell_toolbox_2007}.
For a continuous dynamic $\dot{x} = h(x,u)$, the robust backwards reachable set $R$ of the constraint region $K \subset \mathcal{X} = \mathbb{R}^m$ can be computed as the sub-zero level set of:

\begin{align}
    \frac{\partial V}{\partial \tilde{t}}(x,\tilde{t}) =
    - \min\{0,\max_u \nabla V(x,\tilde{t})^T h(x,u)\}
\end{align}
where $V$ is initialized to the signed distance from $K$:

$$
V(x,0) = dist_K(x)
$$

Let $\Omega_{\tilde{t}}$ be the complement of this backwards reachable set:

\begin{align}
    \Omega_{\tilde{t}} = \{x | V(x,\tilde{t}) \geq 0\}
    \label{eq:safeSet}
\end{align}

As the complement of the reachable set, $\Omega_{\tilde{t}}$ is the set from which there is a way to avoid the keepout set $K$.
Since there exists an avoidant strategy, this $\Omega_{\tilde{t}}$ is a control-invariant set. So long as the system is initialized within $\Omega_{\tilde{t}}$ it is possible to remain safe.
Furthermore, any controls can be taken up to crossing the border from $\Omega_{\tilde{t}}$ into $R$. At this point, the maximally safe action $u(x,\tilde{t})^* = \arg\max_u  \nabla V(x,\tilde{t})^T h(x,u)$ must be taken.
This is the safety strategy advanced in \cite{hoffmann_decentralized_2008}.

This safety strategy ensures the system will stay on the interior of the feasible region.
Due to intervening only when absolutely necessary (i.e. when crossing into $R$), this intervention is also the \emph{least restrictive}. It will not eliminate any trajectories that weren't already infeasible.
Therefore the set of feasible solutions remains unchanged after instituting these dynamics.
The optimal trajectory of the non-intervened dynamics will be the same as the optimal trajectory on the intervened dynamics.

This constraint-enforcing switching control is non-differentiable, so derivative-based optimizations on the controls cannot be used.
Fortunately, new relaxations of switched dynamics \cite{westenbroek_new_2018} can substitute a relaxed problem whose solutions will converge to the true unrelaxed solution as the relaxation is tightened.

\subsection{Constraint Inference Performance} 
Accurate constraint inference relies on a close match between expert demonstrations and soft-optimal trajectories on the tabular MDP that incorporates the ground-truth constraint. For the purposes of constraint inference, two trajectories are equivalent if they violate the same constraints in the constraint hypothesis set. Therefore, we next examine the difference between the expected constraint violation under the tabular MDP and the actual constraints violated by $N=65$ independent continuous demonstrations. Over all of the possible constraints, this difference can be expressed as 

\begin{equation}
    \sum_{i=1}^{100} | \Phi_{i,T} - \frac{1}{N}\sum_{j=1}^{N} D^{(j)}_i|
    \label{eqn:distribution_distance}
\end{equation}

\begin{figure}
    \includegraphics[scale=0.33]{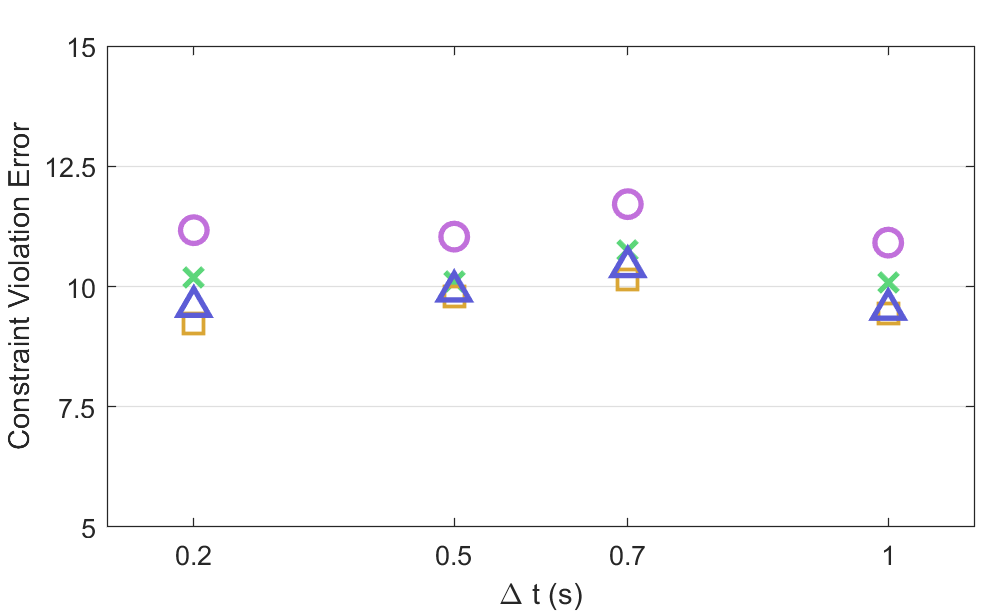}
    \caption{Accuracy in estimating constraint violation for random combinations of start and goal is relatively constant over various hyperparameter values, although there is a small benefit for increasing the state space grid size from 100 cells (pink circles) to 400 cells (green crosses). The metric used here is quantified in equation (\ref{eqn:distribution_distance}). Pink circles: 100 state cells, green crosses: 400, yellow squares: 900, blue triangles: 1600.}
    \label{fig:distribution}
\end{figure}

Where $D^{(j)}_i$ is an indicator for whether demonstration $j$ violates constraint $C_i$. If the approximate MDP perfectly tracks the true constraint violation distribution and demonstrations are distributed according to soft optimality, we expect the quantity in equation (\ref{eqn:distribution_distance}) to go asymptotically to 0 as the number of demonstrations increases. Results for different model hyperparameters and a single demonstration (averaged over 65 trials and the two alternative ground-truth constraints) are shown in Fig. \ref{fig:distribution}. Increasing the number of state space grid cells from 100 to 400 lowers constraint violation error, but increasing the number of states in the tabular MDP beyond this point does not have much effect. This suggests that the greater accuracy of the approximate MDP for larger numbers of discrete states does not necessarily translate into improved constraint inference. Error is stable across different values of the discrete time interval $\Delta t$. 

\begin{figure}
    \includegraphics[scale=0.34]{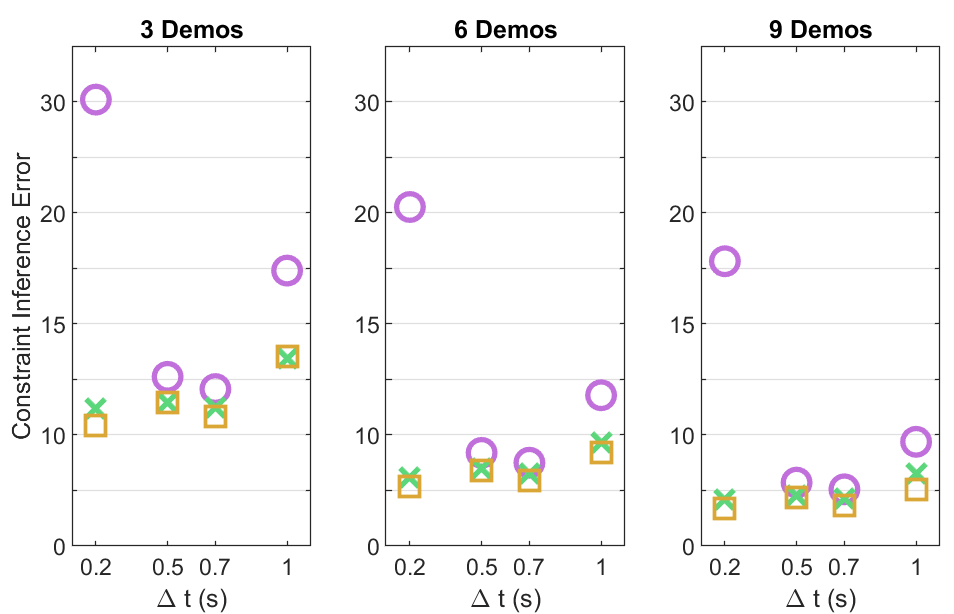}
    \caption{The true constraint is ranked as the one of the most likely possibilities after just a few demonstrations across many hyperparameter choices. For the 10x10 state space grid (pink circles), $\Delta t = 0.2$ gives poor performance because it often isn't possible for the agent to reach a new cell within this time frame, so most transitions result in the tabular agent erroneously staying in the same location. Overall, there is a small benefit for increasing from 100 state cells to 400 cells (green crosses). A 1600 state cell version performs equivalently to 900 cells (yellow squares) but is omitted for clarity. Pink circles: 100 state cells, green crosses: 400, yellow squares: 900.}
    \label{fig:ranking}
\end{figure}

We see a very similar trend when examining the performance of constraint inference across MDP's generated with different hyperparameters, as can be seen in Fig. \ref{fig:ranking}. After choosing an appropriate $\Delta t$, tabular MDP's with 100 to 1600 states are able to successfully identify the true constraint as one of the top-5 likeliest constraints after 9 demonstrations. Increasing the number of states to at least 400 stabilizes performance across different choices of $\Delta t$. Even though the approximate MDP's do not capture the true continuous dynamics with high fidelity, especially for the coarsest state-space grid, constraint inference still works well and is robust to a range of hyperparameters. 

\begin{figure*}
    \includegraphics[scale=0.38]{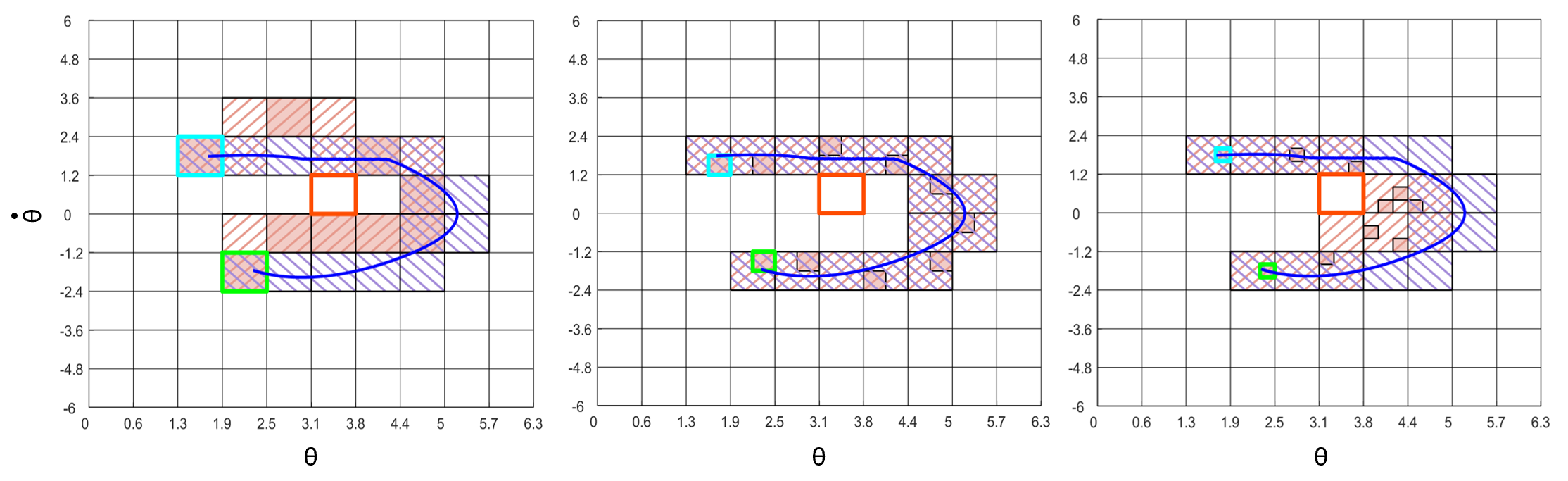}
    \caption{We can directly compare the approximated MDP to individual demonstrated trajectories by sampling a discrete trajectory from the MDP when the true constraint is known. From left to right, a trajectory sampled from an MDP with 100, 900, and 1600 states, respectively. The blue line is the demonstration, which is attempting to get from the start (cyan box) to the goal (green box) while avoiding the true constraint (red box). The states in the discrete trajectory are shown with shaded pink boxes. Possible constraints violated by the demonstration are shown with blue diagonal lines going from top left to bottom right, while possible constraints violated by the discrete trajectory are shown with pink diagonal lines going from bottom left to top right. For the finer grid state spaces in the center and right panels, the states visited by the discrete trajectory are smaller than the constraint regions. For this example, the trajectory from the 900-state MDP happens to most closely match the demonstration. }
    \label{fig:trajectory}
\end{figure*}

In addition to these average trends, we can qualitatively examine the approximation quality by sampling a trajectory from the discrete MDP and comparing it to the original continuous demonstration. An example of this for a single trial is shown in Fig. \ref{fig:trajectory}.

For all of the analyses described above, we also varied the number of discrete actions in the approximate MDP but found that this made little difference to any of the measures we examined. A larger number of actions allows the discrete agent more possible routes to the goal, but it may be that these routes do not change constraint violation behavior in expectation across the soft-optimal policy distribution. Fig. \ref{fig:accuracy} through Fig. \ref{fig:trajectory} show results using 9 actions evenly spaced from $u=-2$ to $u=2$.

\subsection{Confidence In Found Constraints}
In addition to identifying the most likely constraints influencing agent behavior, it is desirable to calculate the probability of there being a constraint at all. First, consider the simple case where we assume that there is at most one constraint, and if there is one, it is the most likely one identified via MLCI. Let $C$ be the event that this is truly a constraint, and $A_N$ be the event that N independent trajectories do not violate this constraint. We would like to calculate $P(C|A_N)$. We know that $P(A_N|C) = 0$ since no demonstrations may violate a constraint, and that $P(\bar{A_1}|\bar{C})=\Phi_T$ (i.e. the probability of a demonstration not violating this constraint by coincidence, even if the agent isn't really subject to it), which we obtain from the MLCI algorithm. We can therefore use Bayes' Rule to obtain the following formula:

\begin{equation}
    P(C|A_N) = \frac{P(C)}{P(C)-(1-\Phi_T)^{N}(1-P(C))}
\end{equation}

where $P(C)$ is a prior on the probability of the constraint being present. This simple formula introduces no additional approximation error beyond what is already present in the model under the assumptions described above, and presents an important advantage of the MLCI approach to constraint inference over previous approaches that cannot provide confidence estimates of found constraints. Unfortunately, relaxing the assumptions on possible constraints and calculating probabilities of all possible constraints quickly becomes computationally intractable. Providing estimates of these probabilities is left for future work.

\section{Potential Application: Sit-to-Stand and Lower Back Pain} 
The robustness to hyperparameters selection, low number of required demonstrations, and ability to provide a confidence interval on the identified constraints supports the use of the MLCI approach to identify patient-specific impairments from observed motion. One potential application is in the analysis of individuals with Low back pain (LBP). 

LBP affects 70-90\% of adults during their lifetime and can be extremely debilitating \cite{thiruganasambandamoorthy_risk_2014}. However, it is often difficult to determine the source of the pain and therefore prescribe an appropriate treatment. Disorders of the lower spine, hip, and pelvic region can all cause LBP \cite{prather_links_2019}. Treating the wrong problem may result in an unnecessary surgery that doesn't resolve the patient's LBP. When a treatment plan that addresses the physical cause of the pain can't be identified, patients may be prescribed opioids for chronic pain management, even though these are ineffective and can lead to abuse and addiction \cite{martell_systematic_2007}. Therefore, there is a pressing clinical need to develop better methods for understanding the source of LBP.

There is a recent body of literature suggesting that LBP may be linked to irregularities in movement patterns. For example, inappropriate amounts of pelvic movement during various motions appears to contribute to LBP \cite{m_correlation_2015}. This pelvic movement may be compensatory for a limited range of motion in other joints - in other words, constraints on the achievable joint angles. We would expect the resulting movement patterns to avoid regions of the biomechanical state space associated with pain. Identifying both physical and pain-related constraints on movement could therefore lead us to a better understanding of the underlying cause of the LBP. For this reason, we would like to infer the most likely constraints a person is acting under when observing their movements. A particularly promising movement pattern for demonstrations is completing a sit-to-stand trajectory, which exerts significant strain on several joints implicated in LBP \cite{hughes_chair_1994}. A telescoping inverted pendulum system has been used to model this movement, which reduces the problem to 4 dimensions while allowing for clinically relevant discovery \cite{papa_telescopic_1999}.

\subsection{Constraint Inference on Telescoping Inverted Pendulum}

Following the above motivation, we next demonstrate successful constraint inference on a telescoping inverted pendulum (TIP) model. The dynamics for this model are as follows:

\begin{align*}
    \ddot{\theta} &= \frac{g}{l} \cdot sin(\theta) + u_1 \\
    \ddot{l} &= u_2
\end{align*}

\begin{figure}
    \centering
    \includegraphics[scale=0.3]{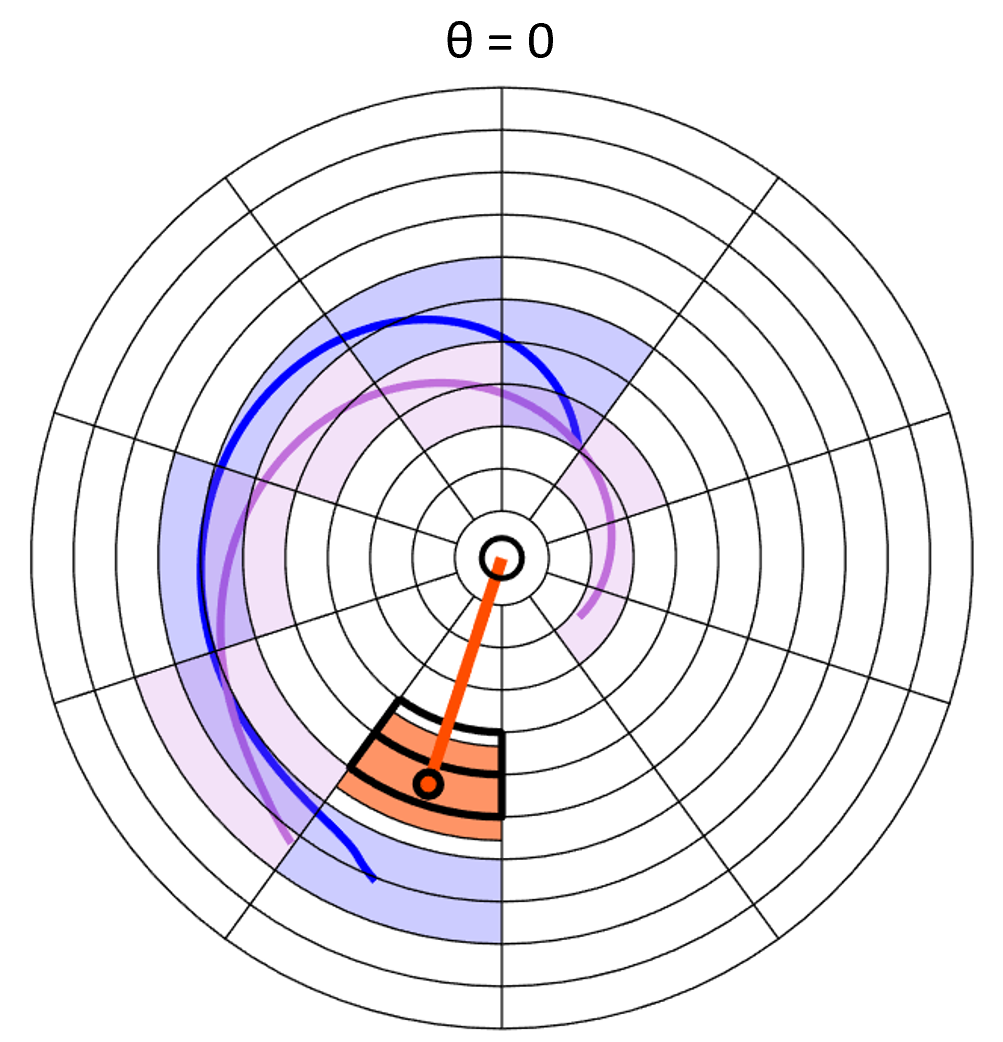}
    \caption{Ground truth and inferred constraints for a 4-dimensional telescopic inverted pendulum (TIP) model. The constraint hypothesis space is 100 evenly spaced cells in length and angle. The ground truth constraint, which does not evenly align with the constraint hypothesis set, is shown in red. After 5 demonstrations, the top 2 most likely constraints, shown as cells with dark black outlines, coincide very well with the true constraint. 2 demonstrations of those used in inference, along with the constraints they violated (shaded cells), are shown in blue and purple. The tabular MDP must keep track of the linear and angular velocity dimensions as well as the position dimensions shown here.}
    \label{fig:tip}
\end{figure}

These dynamics omit the cross-coupling term between angular acceleration and linear velocity for simplicity. For this experiment, we chose the goal set as the set of all states within a certain range of pendulum length and angle, leaving velocity as a free parameter. The objective is to reach the goal set at $\tilde{T}$ = 5s while minimizing $||u||_2^2$. The constraint hypothesis space is a 10x10 evenly spaced grid along the angle and length dimensions, so that if a particular (angle, length) combination is constrained, the agent is not allowed to enter that combination at any velocity. We generated 5 demonstrations with random start and goal states following the same procedure as in section V-B. We then formulated a tabular MDP with 2500 states (10 cells each for angle and length, and 5 cells each for angular and linear velocity) and 15 actions (5 discrete torque options and 3 discrete linear force options) and performed constraint inference on the demonstrations. The ground truth constraint and top 2 likeliest inferred constraints are shown in Fig. \ref{fig:tip}. Despite the coarseness of the tabular state-action space and a mismatch between the constraint hypothesis space and the true constraint region, MLCI correctly identifies the ground truth constraint and takes about 5 minutes with no optimization effort on a single CPU core. If the start and goal states of demonstrations are known in advance, as is likely to be the case in a clinical test, this computation can be done ahead of time and inferring constraints after observing the actual demonstration trajectories is virtually instantaneous.
 
\section{Conclusion}
We have presented methodology for forming a tabular MDP approximation of continuous dynamics which can be used for maximum likelihood constraint inference. Although the approximation introduces some error into the estimation, constraint inference works well with pendulum dynamics over a range of hyperparameters, including a small discrete state space. The present approach allows for ranking possible constraints by their likelihood, which is especially useful in applications with significant uncertainty, and uses the maximum entropy framework, which may be an especially good fit for human demonstrators, who tend to act sub-optimally. Future work should characterize the kinds of dynamics for which this approach works well and whether techniques such as variable grid size may allow for higher accuracy and increased computational efficiency.

\bibliographystyle{plainnat}
\bibliography{references}

\end{document}